\documentclass[10pt, a4paper]{article}
\usepackage{lrec2022} % this is the new LREC2022 Style
\usepackage{multibib}
\newcites{languageresource}{Language Resources}
\usepackage{graphicx}
\usepackage{tabularx}
\usepackage{soul}
\usepackage{float}
\usepackage{titlesec}
\titleformat{\section}{\normalfont\large\bfseries\center}{\thesection.}{1em}{}%https://www.overleaf.com/project/61a7a4ee33e45d89c3786ad1
\titleformat{\subsection}{\normalfont\SmallTitleFont\bfseries\raggedright}{\thesubsection.}{1em}{}
\titleformat{\subsubsection}{\normalfont\normalsize\bfseries\raggedright}{\thesubsubsection.}{1em}{}
\renewcommand\thesection{\arabic{section}}
\renewcommand\thesubsection{\thesection.\arabic{subsection}}
\renewcommand\thesubsubsection{\thesubsection.\arabic{subsubsection}}
\usepackage{epstopdf}
\usepackage[utf8]{inputenc}

\usepackage{hyperref}
\usepackage{xstring}

\usepackage{color}

\title{A New Dataset for Topic-Based Paragraph Classification in Genocide-Related Court Transcripts}

\name{Miriam Schirmer, Udo Kruschwitz, Gregor Donabauer} 

\address{University of Regensburg \\
         Regensburg, Germany \\
         \{miriam.schirmer, udo.kruschwitz, gregor.donabauer\}@ur.de\\}

\abstract{
Recent progress in natural language processing has been impressive in many different areas with transformer-based approaches setting new benchmarks for a wide range of applications. This development has also lowered the barriers for people outside the NLP community to tap into the tools and resources applied to a variety of domain-specific applications. The bottleneck however still remains the lack of annotated gold-standard collections as soon as one's research or professional interest falls outside the scope of what is readily available. One such area is genocide-related research (also including the work of experts who have a professional interest in accessing, exploring and searching large-scale document collections on the topic, such as lawyers). We present GTC (Genocide Transcript Corpus), the first annotated corpus of genocide-related court transcripts which serves three purposes: (1) to provide a first reference corpus for the community, (2) to establish benchmark performances (using state-of-the-art transformer-based approaches) for the new classification task of paragraph identification of violence-related witness statements, (3) to explore first steps towards transfer learning within the domain. We consider our contribution to be addressing in particular this year's hot topic on Language Technology for All.
\\ \newline  }

\begin{document}

\maketitleabstract

\section{Introduction}

Information overload has led to a multitude of search applications of which Web search is just one out of many. 
Unlike search for leisure or personal interest there is a vast area of search contexts which are found in a work environment. Professional  search falls into that scope, i.e. search over domain-specific document collections and often with search tasks that are recall-oriented rather than precision-focused 
\cite{Kruschwitz17Searching,Verberne19Information}. Beyond applications where such search effort can directly be measured in financial terms (e.g. in patent search, e-discovery or the compilation of systematic reviews) there are many other fields where these costs are more implicit, e.g. in the area of genocide studies that rely on the analysis of vast quantities of different resources \cite{bachman2020cases,hinton2012critical}.

Looking at the wider picture, searching large text corpora for specific thematic patterns can be very time-consuming and non-trivial, in particular for searchers who do not have a solid foundation in NLP or search technology.
The huge amount of court transcripts of genocide tribunals presents a perfect example: the \textit{International Criminal Tribunal of the Former Yugoslavia (ICTY)} alone provides official transcripts for each of its cases online, leading up to approximately 2.5 million pages of transcripts in total \cite{icty_2016}. Searching for specific content in a text corpus like this usually requires vast amounts of manual research capacity \cite{hoang_schneider_2018}. Tools and approaches to augment this type of search and help limit manual efforts have been developed for a broad range of use cases, e.g. for automating search strategies or text extraction from documents \cite{macfarlane2021,russell-rose_et_al_2021}. However, even with the help of suitable tools, searching for specific text passages in large text corpora generally remains a difficult task, in particular when the search is recall-oriented \cite{Bache2011,kapstein_et_al_2013,noor2015evaluating}. 
%(XXX).

Turning to our own use case, 
the search for specific content in transcripts of genocide tribunals further proves difficult because transcripts are only accessible individually (usually one court day per transcript) and in different formats, depending on the tribunal. So far, no datasets of any kind containing genocide court transcripts have been published. Similarly, no other forms of pre-structured or annotated text data in this field of research exist.

This paper addresses this gap by providing a systematically annotated dataset containing text material from three different genocide tribunals: the \textit{Extraordinary Chambers in the Courts of Cambodia (ECCC)}, the \textit{International Criminal Tribunal for Rwanda (ICTR)}, and the \textit{ICTY}. In addition to compiling the sampled corpus, we provide annotations within the text. More specifically, text passages in which witnesses talk about experienced violence have been annotated, focusing on a core part of each testimony. Given that respective passages on violence often cover crimes that are relevant for the indictment of the accused, such as murder or rape, they are essential for judgement. At the same time, they are not easily classifiable due to their potential ambiguity. 

%and at the same time not easily classifiable as they might also be ambiguous. 

%(see section \ref{dataset} for a more detailed description). 
Since this dataset of textual documents is the first in this area, we hope that it provides a valuable resource for NLP-based genocide research. To foster generalisability and assess transferability of approaches, the corpus contains a sample from different tribunals. %including more than one tribunal, this dataset will also help generalise results and draw comparisons between the individual courts.

We consider the provision of the new resource our main contribution, but we also provide experimental work that
will serve as a benchmark and allow the contextualisation within the broader field. We use fine-tuned BERT models for that purpose. Utilizing the heterogeneous nature of the corpus we also explore transfer learning and report results. 

%While providing a comprehensive dataset, this paper simultaneously examines ways of improving and simplifying wide-range search based on transcript material. Specifically, the dataset will be applied to techniques of identifying text paragraphs in court transcripts based on a specific topic. For this purpose a BERT-based transformer model will be applied to transcripts of genocide tribunals, examining how state of the art transformer-based approaches can be applied to this field.

%In this case, text passages in which witnesses talk about experienced violence will be selected, focusing on a core part of each testimony. Since respective passages on violence often cover crimes that are relevant for the indictment of the accused, such as murder or rape, they are essential for judgment.

Beyond the contribution to the NLP community, it is our hope that the results of this paper will be useful for both scholars and practitioners at international criminal tribunals who need to work through large quantities of transcript material as part of their everyday job. \\

We summarize our contributions as follows: 

\begin{enumerate} 
\item We present GTC, a new reference corpus sampled from different international criminal courts in the context of genocide tribunals. The corpus contains annotations of statements by witnesses about experienced violence.
\item We built state-of-the-art transformer-based classifiers to provide benchmarks for the new classification task of paragraph identification of violence-related witness statements.
\item We provide experimental results for transfer-learning by varying the training and testing data across documents from different tribunals.
\item We make all data as well as code available to the community.
\end{enumerate}

\section{Related Work}

We touch on the three key areas of interest our work falls into, namely resources, professional search, and text classification. The discussion of each of these should simply serve as both a motivation and basic context.

\subsection{Resources}

The importance of publicly available language resources to help develop NLP applications has long been recognized, e.g. \newcite{Calzolari10LREC}, and the domain-specific nature of many problems is what makes respositories such as the \textit{LRE  Map}\footnote{\url{https://lremap.elra.info}} a valuable starting point for many researchers and practitioners. 

%For the specific use case of assisting searchers to identify relevant information in genocide-related court transcripts there simply are no suitable resources available, neither in English nor in any other relevant languages \cite{fidahic2021case}. 

For the specific use case of assisting searchers to identify relevant information in genocide-related court transcripts resources are very limited to non-existent. Of course, transcripts of each tribunal are available through the respective courts' websites -- however, their quality in terms of digitisation (e.g., object character recognition) varies greatly. Specifically for the ICTY, \newcite{fidahic2021case} further criticises that transcripts are only available in certain languages, thus limiting access mainly to English-speaking readers.
Considering that the field of genocide research and studies  \cite{totten2017advancing} is multi-faceted enough to warrant the provision of suitable resources, we see our own contribution as a starting point to fill this gap, even though we limit our work to English transcriptions.

\subsection{Professional Search}
%\subsection{Professional Search and Time Allocation}

Searching through court transcripts can often be framed as an instance of professional search \cite{koster_et_al_2009,russell-rose_et_al_2021}. Professional search describes the process of searching for  information  in a work context which is commonly domain-specific and requires expertise in a specific area. Key features of professional search are  limited time and budget resources, making it desirable to 
provide support that helps classify specific text passages which ultimately could drastically reduce search efforts \cite{russell-rose_et_al_2021}. It should be noted that professional search is very different from other types of search such as  Web search. A common observation is that searches take a lot longer to satisfy a specific information need. For example, \newcite{bullers_et_al_2018} found that librarians spend 26.9 hours on average on systematic reviews that involve searching for specific content, indicating that this task is highly time-consuming. Similarly, \newcite{Greene_Colozzi_et_al_2021} discuss the time-consuming process of researching court transcripts and other relevant sources related to cybercrime. Professional search in court transcripts in general, however, has not been analysed so far. 

Another important aspect dealing with extensive search in large text corpora are human factors. Especially when dealing with time-consuming search in text documents that lasts for hours, fatigue might be an issue that reduces the quality of the search. Additionally, manual search is also more vulnerable to subjectivity, motivating the use of automated search algorithms \cite{Li_et_al_2020}.

%Different tools and algorithms to save time in searching through text have been discussed and much of this will vary depending on the specific search context. 

Different tools and algorithms to save time in searching through text have been discussed -- varying strongly depending on the specific search context. For example, \newcite{macfarlane2021} give a broad overview of different tools for systematic literature reviews, such as tools for text and data extraction or automatic query expansion. While these tools might help review literature more efficiently, the authors note that their use is not widespread. Furthermore, not all of the above-mentioned tools are helpful when it comes to content-based search in text documents. In this context -- when looking for specific content in court transcripts -- tools for enhanced keyword search might prove more useful.

\subsection{Text Classification}
%\subsection{BERT as a Reliable Model for Text Classification}

Topic-based paragraph classification specifically for court transcripts has so far not been discussed in the literature. Nevertheless, extensive research in this area has been done in other fields. As a traditional and fundamental NLP task, text classification covers a wide range of tasks ranging from  category labeling over  sentiment analysis to authorship attribution \cite{jurafsky_martin_2021}. %Each of these tasks can be carried out by different means (for an overview of different text classification methods see \cite{Li_et_al_2020}.
Traditionally effective approaches to supervised machine learning, such as Support Vector Machines  \cite{al2018random,tong2001support,zhang2008text} or K-nearest neighbour  \cite{bijalwan2014knn}, have now largely been replaced by transformer-based approaches \cite{dhar2021text,jurafsky_martin_2021,minaee2021deep}.
Fine-tuning BERT has become the standard baseline in text classification \cite{devlin_et_al_2019}, not just beating traditional machine learning paradigms but also recurrent neural networks (RNNs), convolutional neural networks (CNNs) or other deep neural networks (DNNs) \cite{Li_et_al_2020}.

%A very prominent model used for text classification is Bidirectional Encoder Representations from Transformers (BERT) 
%As a pre-trained open-sourced language model BERT generates contextualized word vectors that help classify specific text fragments. Researchers in various fields have emphasized the high performance of BERT in comparison to other architectures, such as recurrent neural networks (RNNs), convolutional neural networks (CNNs) or other deep neural networks (DNNs) \cite{Li_et_al_2020}.

Example topic areas in which BERT has been utilized effectively in text classification include various forms of sentiment analysis  ranging from  aspect-based sentiment analysis  \cite{sun_et_al_2019} to sentiment analysis on the impact of coronavirus in social life \cite{singh_et_al_2021}, as well as reading comprehension tasks, e.g., \newcite{Xu_et_al_2019}. 

Of specific concern to  our underlying use case is text classification that requires \textit{text segment classification}, commonly found when applied to social media data, such as tweets and comments. For this type of analysis, splitting larger text data into paragraphs limited to a certain number of words has been established as a regular step in the NLP pipeline \cite{Li_et_al_2020}. A very prominent example of using BERT sequence classification is hate speech detection (e.g., \newcite{Mozafari_et_al_2019}, \newcite{Mozafari_et_al_2020},
\newcite{sohn_lee_2019}). By applying BERT to Twitter data, tweets can easily be classified according to whether they contained racism, sexism, or hate, among others \cite{Mozafari_et_al_2019}.

New BERT models and applications are being reported at rapid speed as the model is continuously applied in new fields. Examples are the recently developed ClimateBert, a pre-trained language model for climate-related text \cite{webersinke_et_al_2021} or COVID-Twitter-BERT (CT-BERT) \cite{Mueller_et_al_2020}.

\subsection{Concluding Remarks}
%\subsection{BERT as a Reliable Model for Text Classification}

The new corpus we provide aims to bridge a (domain-specific) gap that exists in the landscape of annotated text collections. In order to assess the utility of the corpus and the difficulty of the underlying classification task we will adopt the commonly applied baseline approach of fine-tuning BERT. 
One of the goals is to show whether or not BERT also serves as an efficient tool for this type of text data and whether it can help simplify classification of paragraphs in court data.

This can only be a first step at filling the identified gap -- there will be scope for many future directions, not least to replicate the approach to other languages.

\section{Genocide Transcript Corpus (GTC)}
\label{dataset}

We introduce \textit{Genocide Transcript Corpus (GTC)}, a corpus of transcripts drawn from the court proceedings of international tribunals dealing with cases of genocide. Following sampling of the original data we also apply an annotation step that assigns binary labels to individual paragraphs.\footnote{The full dataset and code are  accessible online: \url{https://github.com/MiriamSchirmer/genocide-transcript-corpus}.} The paragraph labeling is aimed at identifying those parts of the text that refer to violence experienced by witnesses --  relevant are only those text segments which are actually part of witness statements.

The dataset used in this study consists of 1475 text passages from three different genocide tribunals. Transcripts from the three biggest ad-hoc genocide tribunals, the ECCC, the ICTR, and the ICTY were selected. In a first step, the courts' databases were searched for witnesses who have actually experienced some form of violence. This pre-selection ensured having a substantial amount of relevant text passages in the dataset and thereby excluding technical or expert witnesses. Three different tribunals were selected to provide a diverse dataset and explore transfer learning, i.e. to show possible differences in the results after training and testing with data from different tribunals. Thus, results are more generalisable and differences in individual tribunals are controlled for.

Between 4 and 7 transcripts were selected per tribunal and were divided into equally large text chunks of 250 words each. Numbers and punctuation were removed in a first preprocessing step. In the final dataset, the number of samples is roughly equally distributed across  tribunals (ECCC: 465, ICTY: 530, ICTR: 480). Differences occur since only complete transcripts with varying length (about 40 to 120 pages) were included. \\

The current version of the GTC contains the following data:

\begin{itemize}
\item For the ECCC, transcripts with a total of 438 pages from two different trials (Case 001 against Kaing Guev Eav, Case 002 against Nuon Chea and Khieu Samphan) were selected. This includes the proceedings of 4 full court days and the hearing of 7 witnesses. 

\item Transcripts of the ICTY were taken from the cases against Slobodan Milošević (IT-02-54) and Duško Tadić (IT-94-1). The material consists of 416 pages of transcripts from 5 trial days, with 15 witnesses testifying in court. 

\item For the ICTR, 566 pages of transcript material from the cases against Jean-Paul Akayesu (ICTR-96-04) and Pauline Nyiramasuhuko et al. (ICTR-98-42) were included in the dataset. The ICTR data includes 5 witnesses and 7 court days. 
\end{itemize}

In total, 1420 pages of transcripts were incorporated into the dataset. Differences in the number of pages and witnesses are firstly due to different transcript formats regarding digitisation  and text density per page. Secondly, legal proceedings vary between the different tribunals and thus lead to slightly different content. For example, in the selected ECCC and ICTR transcripts, witnesses are questioned for approximately one court day, whereas in the selected ICTY transcripts, 2 to 3 witnesses were questioned per day.

\section{Methodology}

\subsection{Label Annotation}

\begin{figure}[!ht]
    \centering
    \includegraphics[width=\columnwidth]{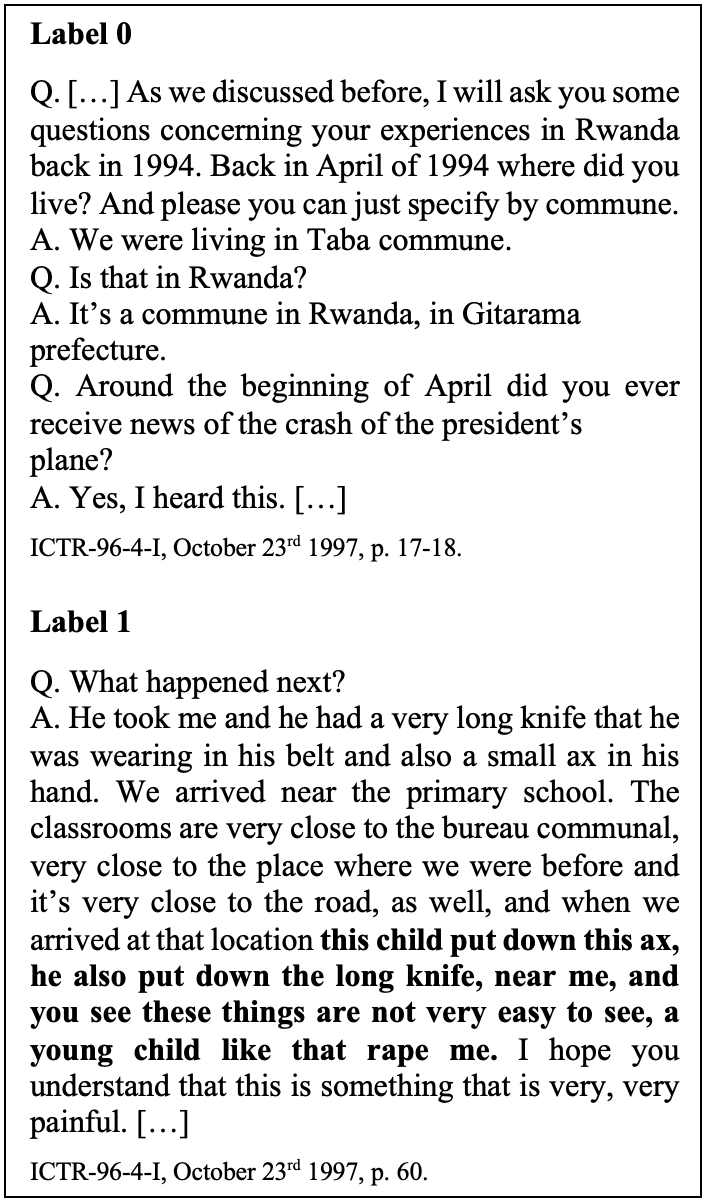}
    \caption{Sample abstracts from the corpus demonstrating two clear-cut examples for  a text passage that does \textit{not} contain accounts of violence in a witness statement (top example -- Label 0) and one that does (bottom example - Label 1). The examples were shortened, and both format and punctuation were adapted for readability.}
    \label{fig:fig1}
\end{figure}

All samples were labeled according to whether they contain a witness’s description of experienced violence (0 = no violence, 1 = violence). Violence in this context is interpreted broadly and includes accounts of experienced or directly witnessed torture, interrogation, death, beating, psychological violence, experienced military attacks, destruction of villages, looting, and forced displacement. We restrict our interest to a binary classification, i.e. different acts of violence were not categorized further into subcategories. Figure \ref{fig:fig1} provides an example of a rather clear distinction between the two labels.

\begin{figure}[!ht]
    \centering
    \includegraphics[width=\columnwidth]{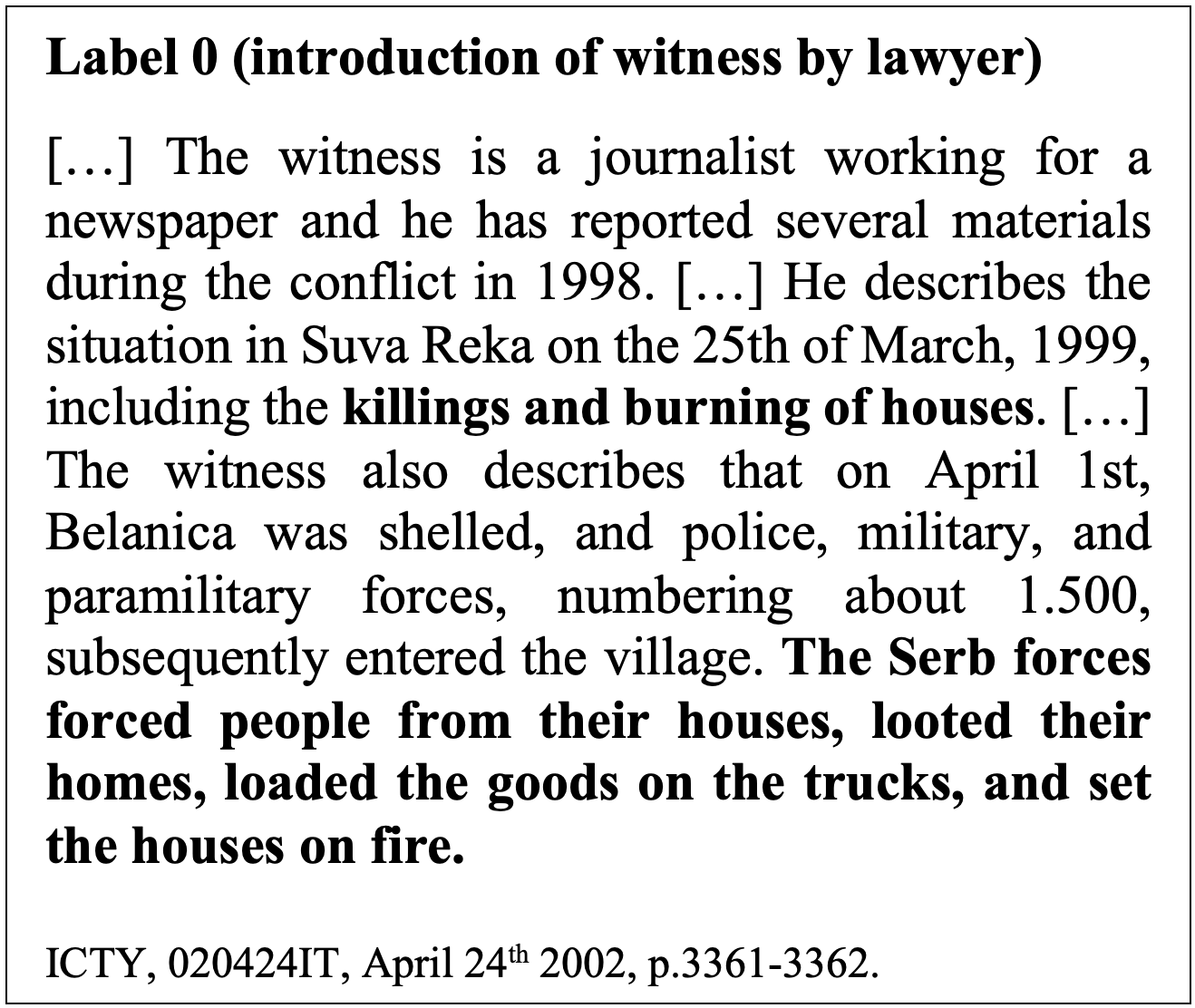}
    \caption{Example of a text passage that contains violence-related vocabulary, but is not labeled 1. As in Fig. 1, this example was shortened and adapted.}
    \label{fig:fig2}
\end{figure}

 An important requirement for labeling text passages as containing accounts of violence was whether experienced violence was described by the witness orally in court. Questions by lawyers and judges containing violence-related words were thus labeled '0'. However, since the words used in both cases are the same for the most part, the differentiation between violence-related statements of witnesses vs. lawyers, judges, or the accused makes an automated classification more difficult. Having written statements (e.g., statements recorded previously by court staff, police, or human rights organisations) read out loud during the trial increases this difficulty further: even though reports contain accounts of experienced violence, they are not labeled '1' because they were not expressed orally by the witnesses during the trial, but by a lawyer or another representative of the court (see Figure \ref{fig:fig2} for an example).
 
It should have become clear that the task of correct classification in an automated fashion is non-trivial; simple 'bag of words'-based approaches are likely to underperform. Apart from the context that makes it clear how to classify a paragraph, looking at the vocabulary alone will not be sufficient. A similar observation was made when classifying a corpus of tweets which were classed as falling into a number of different classes all to do with violence such as crises, violence, accidents, and crime \cite{alhelbawy_et_al_2016}. It  was found that the inter-rater agreement varied significantly across the different violence classes.

%That the detection of violent content can prove difficult has been shown in a dataset of Arabic tweets that contain some form of violent event \cite{alhelbawy_et_al_2016}: Tweets were categorized into different violence classes, such as crises, violence, accidents, and crime. Having up to 10 contributors annotating the individual tweets, the authors found that the interrater agreement varied significantly across the different violence classes.

Since this dataset does not differentiate between subcategories, classification was limited to a binary task. However, to make sure that the categorization is reliable, a random selection of approximately 200  samples were independently labeled by a second researcher (with an inter-rater reliability $\kappa$ = 0.86) according to the above-mentioned facets of experienced violence. Table \ref{table:number_of_samples} provides an overview of the number of each label per tribunal. Differences in the label balance are due to the random selection of transcripts.

\begin{table}[!ht]
\begin{center}
\begin{tabularx}{\columnwidth}{X|X X X}
      \hline
       & n$_{0}$ & n$_{1}$ & n$_{total}$\\
      \hline
      ALL & 946 & 529 & 1475\\
      \hline
      ECCC & 286 & 179 & 465\\
      \hline
      ICTY & 401 & 129 & 530\\
      \hline
      ICTR & 259 & 221 & 480\\
      \hline
\end{tabularx}
\caption{Overview of label balance for the complete dataset (“ALL”) and the three individual tribunal datasets.}
\label{table:number_of_samples}
\end{center}
\end{table}

\subsection{Experimental Setup}

For all experiments, the 12-layer BERT$_{base}$ architecture for sequence classification \cite{devlin_et_al_2019} was used to classify text passages of genocide tribunal transcripts.

As described in Section \ref{dataset}, the dataset consists of 3 subsets with data from different tribunals. 5-fold cross-validation (80:10:10) was applied to each subset and to the full version of the dataset (concatenated subsets).

Overall, BERT$_{base}$ was trained on all possible train, validate and test constellations, leading to a total of 16 different combinations. In those cases, in which training, validation and test data originate from the same subset, the respective splits led exactly to an 80:10:10 distributed number of samples. When training on one (or more) class(es) and testing on samples of a single remaining class, we held out all samples of the target class for testing. Consequently, for some of the combinations the number of test samples equals or even exceeds the number of samples in the train and validation data (for details see Table \ref{table:dataset_combination}).

In a first step, BERT$_{base}$ was trained on the full dataset to classify samples of all three tribunals, but also to classify tribunal-specific text chunks.

Secondly, we apply the same setup to all three subsets. More specifically, training was performed using tribunal-specific samples to see if BERT is still able to predict class labels of both, the mixed dataset (excluding training class), as well as the remaining tribunal-specific subsets.

To test for the detection of undersampled violence-related paragraphs, additional experiments on this data were set up. All of the subset-specific negative class samples were used and a random proportion of 20\% of positive class samples was added.
 
 %As the size of subsets is limited to 214-288 paragraphs, the development and test sets most of the time solely included a single positive class sample, making it impracticable to test this experimental setup.

For training and validation a batch-size of 16 samples and an epoch-number of 3 (compare \newcite{devlin_et_al_2019}) was used. The training was executed using 4 Nvidia RTX 2080Ti GPUs with an overall memory size of 44GB.

Precision, recall, micro and macro F1 scores for each train/validate/test constellation are provided -- in line with common practice, macro F1 scores will be the reference score when comparing results  \cite{jurafsky_martin_2021}.

\begin{table*}[ht!]
\begin{center}
\begin{tabularx}{\textwidth}{l|X|X|X|X}
      \hline
      \textbf{Train/val data} & \multicolumn{4}{|c|}{\textbf{Test data}} \\
      \hline
      Mixed dataset & Mixed dataset (n$_{train}$=1180, n$_{val}$=147, n$_{test}$=148) & ECCC dataset (n$_{train}$=808, n$_{val}$=202, n$_{test}$=465) & ICTY dataset (n$_{train}$=756, n$_{val}$=189, n$_{test}$=530) & ICTR dataset (n$_{train}$=796, n$_{val}$=199, n$_{test}$=480)\\
      \hline
      ECCC dataset & Mixed dataset (n$_{train}$=372, n$_{val}$=93, n$_{test}$=1010) & ECCC dataset (n$_{train}$=372, n$_{val}$=46, n$_{test}$=47) & ICTY dataset (n$_{train}$=372, n$_{val}$=93, n$_{test}$=530) & ICTR dataset (n$_{train}$=372, n$_{val}$=93, n$_{test}$=480)\\
      \hline
      ICTY dataset & Mixed dataset (n$_{train}$=424, n$_{val}$=106, n$_{test}$=945) & ECCC dataset (n$_{train}$=424, n$_{val}$=106, n$_{test}$=465) & ICTY dataset (n$_{train}$=424, n$_{val}$=53, n$_{test}$=53) & ICTR dataset (n$_{train}$=424, n$_{val}$=106, n$_{test}$=480)\\
      \hline
      ICTR dataset & Mixed dataset (n$_{train}$=484, n$_{val}$=96, n$_{test}$=995) & ECCC dataset (n$_{train}$=384, n$_{val}$=96, n$_{test}$=465) & ICTY dataset (n$_{train}$=484, n$_{val}$=96, n$_{test}$=530) & ICTR dataset (n$_{train}$=484, n$_{val}$=48, n$_{test}$=48)\\
      \hline
\end{tabularx}
\caption{Overview of sample balance for the complete, mixed dataset (“ALL”) and the three individual tribunal datasets for each train/validate/test constellation.}
\label{table:dataset_combination}
\end{center}
\end{table*}

\section{Results}
\label{results}

Our results show that a binary classification based on BERT yields very reliable results across text data from different tribunals. A macro F1 score of 0.81 when training, testing and validating with the complete, mixed dataset that includes all three tribunals shows that BERT can be applied to this type of data and provides reasonably good predictions across the different subsets.

Considering the individual tribunals, using a tribunal-specific dataset for training and validating provided varying test results (ECCC-ECCC macro F1=0.70; ICTY-ICTY macro F1=0.68; ICTR-ICTR macro F1=0.80). Overall, using the mixed dataset for training and validating resulted in the highest F1 scores throughout the tribunal variations (min macro F1=0.78, max macro F1=0.85), independently of the dataset that was used for testing. The highest individual F1 score in our experiments was obtained when predicting data from ICTR transcripts with trained and validated data from the mixed dataset ("ALL")
% this is wrong, right? only true for micro-F1: or ICTR only 
(macro F1=0.85). 

Looking at the tribunal-specific outcomes for the respective training/validating/sets also yielded solid results overall: 
%Training with the mixed dataset ("ALL") was most successful when testing with mixed data from all three tribunals (macro F1=0.81) or data specifically from ICTR transcripts (F1=0.85). 
Interestingly, using the ECCC data for training and validating has the highest true prediction rates when testing is conducted with ICTR data (macro F1=0.79), whereas using ECCC data for training, validating \textit{and} testing only led to a comparatively low macro F1 score of 0.70. When training with ICTY data, performance was also best when predicting ICTR data (macro F1=0.81). Results are similar for training and validating with ICTR data: The highest macro F1 score (0.80) was obtained when using ICTR data for testing.

Overall, precision and recall turned out to be fairly balanced throughout the different training and testing processes. See Table \ref{table:results} for a detailed overview of the results.

%with precision rates being slightly higher on average. For more detailed

When conducting the experiments with undersampled violence-related data, results turn out to be different. Despite using class weights for training (due to the underrepresented positive label), the results obtained are much lower than those reported for the full dataset. For each subset (ECCC: macro F1=0.51, micro F1=0.81; ICTY: macro F1=0.45, micro F1=0.74; ICTR: macro F1=0.45, micro F1=0.75) as well as for the mixed dataset (macro F1=0.47, micro F1=0.77) macro F1 scores are about half of the values reported so far. Since positive samples are heavily underrepresented (e.g. 1 out of 31 samples in the test set) precision, recall and binary F1 for this class amount to 0.0 for a range of data splits. This leads to the overall poor results for this setup. It also offers directions for future experiments.

\begin{table*}[ht!]
\begin{center}
\begin{tabularx}{\textwidth}{l||X|X|X|X||X|X|X|X||X|X|X|X||X|X|X|X}
      \hline
        & \multicolumn{4}{c||}{\textbf{ALL}} & \multicolumn{4}{c||}{\textbf{ECCC}} & \multicolumn{4}{c||}{\textbf{ICTY}} &
        \multicolumn{4}{c}{\textbf{ICTR}}\\
      \hline
       & P & R & mac. F1 & mic. F1 & P & R & mac. F1 & mic. F1 & P & R & mac. F1 & mic. F1 & P & R & mac. F1 & mic. F1\\
      \hline
      \textbf{ALL} & 0.81 & 0.83 & 0.81 & 0.83 & 0.81 & 0.82 & 0.82 & 0.82 & 0.78 & 0.78 & 0.78 & 0.83 & 0.85 & 0.85 & 0.85 & 0.85\\
      \hline
      \textbf{ECCC} & 0.77 & 0.77 & 0.77 & 0.79 & 0.77 & 0.72 & 0.70 & 0.75 & 0.73 & 0.71 & 0.71 & 0.78 & 0.81 & 0.79 & 0.79 & 0.80\\
      \hline
      \textbf{ICTY} & 0.77 & 0.78 & 0.77 & 0.78 & 0.77 & 0.78 & 0.77 & 0.78 & 0.70 & 0.73 & 0.68 & 0.74 & 0.81 & 0.81 & 0.81 & 0.81\\
      \hline
      \textbf{ICTR} & 0.74 & 0.74 & 0.74 & 0.78 & 0.79 & 0.77 & 0.78 & 0.79 & 0.69 & 0.74 & 0.70 & 0.75 & 0.83 & 0.78 & 0.80 & 0.85\\
      \hline
\end{tabularx}
\caption{Results for macro precision (P), macro recall (R) and macro/micro F1 scores on test data (columns) with respect to different training/evaluation set (rows) combinations.}
\label{table:results}
\end{center}
\end{table*}

\section{Discussion}

\textbf{General Discussion:} This study presented a new type of dataset for NLP-based research in the field of genocide and violence studies. BERT$_{base}$ was further used to predict if text passages from court transcripts of three different genocide tribunals contain accounts of experienced violence by the respective witnesses.

The results, in line with expectations, indicate that the mixed dataset is most successful when predicting if a certain text passage from one of three genocide tribunals contains accounts of experienced violence by a witness. Even when classifying paragraphs of one specific tribunal (e.g., the ECCC) with the model that was trained with data from the same tribunal (ECCC in this case), the model trained on the complete dataset provides better results. Including additional data from other tribunals thus improves the quality of the classification.

\textbf{Contextualisation:} Looking at the wider picture, binary classification scores vary widely across NLP applications \cite{arase2019transfer,wang2018glue} -- direct comparisons with other studies must therefore be interpreted with caution. Nonetheless, the ballpark figures we obtained are comparable to state-of-the-art (BERT-based) performance on some other commonly used binary classifications such as MRPC \cite{zhang2020revisiting}, but fall short of performance levels expected for settings \cite{d2020bert}. On the one hand, this confirms once more that BERT can be successfully applied to our corpus and perfectly presents how well this language model has been developed throughout the recent years. On the other hand, further fine-tuning will be necessary to solve performance-related shortcomings.

%Comparing the outcome with previous findings from similar studies, the results of this study remain promising (???). For example, when applying BERT to the General Language Understanding Evaluation (GLUE) benchmark \cite{wang2018glue} to perform a binary single-sentence classification task, \newcite{devlin_et_al_2019} found XXX.

% \textbf{XXX}
% Eventuell könnte man auch das etwas aktuellere SuperGLUE Paper nehmen?
% Weitere Studien für Vergleichswerte: auch mit CoLa/GLUE \cite{zhang2020revisiting} \cite{arase2019transfer}, BERT und Hatespeech \cite{d2020bert}.
% War mir unsicher, was genau die Scores sind, die wir brauchen - aber dachte ich kopiere immerhin schonmal die Paper rein und du siehst das dann vermutlich schneller.
% \textbf{XXX}

% Since BERT performs slightly worse with the genocide transcript data than with other classification tasks, further fine-tuning might be necessary -- especially with this type of data that has not been widely used for NLP tasks.

\textbf{Precision vs. Recall:} The overall similarity of precision and recall rates in our dataset implies that this type of classification might be useful for a broad range of applications. In some cases, recall rates might be more important than precision rates: for example, similar to patent search \cite{Bache2011,bashir2010retrievability}, a high recall is especially important when avoiding missed positive classifications is crucial. In a genocide-transcript-related context, this could apply to staff members who have to work through court transcripts as part of their daily work routine, e.g. for preparing a case. For this option specifically, applying the classification algorithm reduces the time spent on manual search drastically, making sure that no sample is missing and leaving time for manual adaptions. On the other hand, in the context of fast and efficient search with less time for manual adaptions, high precision rates would be more useful, e.g., when only some examples of relevant text segments are required and correctness is more important than completeness \cite{kong_allan_2016}. 

\textbf{Number of Text Chunks and Labeling Balance:} When looking at the results of this study, imbalances in the number of text chunks, labels and in the train-validate-test-ratio must be kept in mind. Still, in spite of the ICTY data containing less violence-related text segments, results did only differ slightly, indicating that this label imbalance does not impact the results significantly.
%That ICTY data for training provides the lowest F1 scores throughout the combinations might be explained by the fact that the ICTY text chunks contained less violence than text paragraphs from other tribunals.

However, extending the dataset further would be a first step in making the results more stable. More text data could also help to improve the label balance: by selecting more transcripts per tribunal, the chances of choosing transcripts that contain no/few or above-average accounts of violence can be reduced.

An extended dataset would also make it easier to experiment with undersampled violence-related paragraphs. As already mentioned in section \ref{results}, this setup currently lacks a sufficient number of positive labels in the test sets (when undersampling this class in an adequate ratio to keep the overall number of samples stable). Thus, adding more (non-violent) text chunks would make it easier to generate representative training/validation/test splits with a sufficient number of paragraphs for both classes regarding this setup. However, the dataset as it is offers directions for a range of possible experiments including the identification of violence related text chunks when heavily underrepresented.

\textbf{Future Research:} This dataset has the potential of serving as a basis for a variety of research approaches in the field of genocide research in the future. For example, more in-depth comparisons between linguistic or content-based characteristics between the three tribunals could be made, building bridges between the interdisciplinary field of genocide research and NLP-approaches. From a NLP perspective, next steps could include further fine-tuning of BERT and establishing a model version that is pre-trained specifically on court transcripts of genocide tribunals. Not least we see our work as a first step towards a downstream practical search system.

\section{Conclusion}

This paper introduces a new dataset of genocide transcript data as a basis for further NLP research and applications. In addition, a baseline for classifying the transcript samples into violent or non-violent text chunks respectively is provided. The results, based on the state of the art BERT architecture, demonstrate that such models can successfully be applied to this new domain and its related classification task. Although the number of text segments used in this study could be further extended (as it especially was observable during experiments with undersampled violence related paragraphs), classification with BERT proved to be successful, emphasizing once more the potential this language model holds even for research areas that have not been in the focus of NLP applications.

\section{Ethical Considerations}

All of the transcripts used in this paper are published online on the respective courts' websites and are publicly accessible. Since this type of text material contains personal and highly sensitive information about witnesses before international criminal courts, special care was taken to ensure that text fragments were not taken out of context. The use of witness names (or their anonymisation) in the dataset was adopted according to the original court document.

\section{Acknowledgements}

To be added. 
% Place all acknowledgements (including those concerning research grants and funding) in a separate section at the end of the paper.

% \nocite{*}
\section{Bibliographical References}\label{reference}
%\label{main:ref}

\bibliographystyle{lrec2022-bib}
\bibliography{lrec2022-example}

%\section{Language Resource References}
%\label{lr:ref}
%\bibliographystylelanguageresource{references}
%\bibliographylanguageresource{languageresource}

\end{document}